\documentclass{article}
\usepackage{spconf,amsmath,graphicx,hyperref}

\usepackage{amsmath}
\usepackage{amssymb}
\usepackage{multirow}
\usepackage{booktabs}
\usepackage{colortbl}
\usepackage{xcolor}
\usepackage{array}
\usepackage{pifont}
\usepackage{bm}
\usepackage[misc]{ifsym}

\definecolor{mycolor}{RGB}{241,240,255}

\title{CP Loss: Channel-wise Perceptual Loss for Time Series Forecasting}
\name{Yaohua Zha$^{1,2}$\ \  Chunlin Fan$^{1,*}$\ \  Peiyuan Liu$^1$\ \  Yong Jiang$^{1}$\ \ Tao Dai$^{3}$\ \   Hai Wu$^{2}$\ \  Shu-Tao Xia$^{1,2}$
\thanks{$^*$Corresponding author.}}
\address{
$^1$Tsinghua Shenzhen International Graduate School, Tsinghua University \\
$^2$Institute of Perceptual Intelligence, Pengcheng Laboratory \\
$^3$College of Computer Science and Software Engineering, Shenzhen University \\
}
\begin{document}
%
\maketitle
\begin{abstract}
Multi-channel time-series data, prevalent across diverse applications, is characterized by significant heterogeneity in its different channels. However, existing forecasting models are typically guided by channel-agnostic loss functions like MSE, which apply a uniform metric across all channels. This often leads to fail to capture channel-specific dynamics such as sharp fluctuations or trend shifts. To address this, we propose a \textbf{C}hannel-wise \textbf{P}erceptual \textbf{Loss} (CP Loss). Its core idea is to learn a unique perceptual space for each channel that is adapted to its characteristics, and to compute the loss within this space. Specifically, we first design a learnable channel-wise filter that decomposes the raw signal into disentangled multi-scale representations, which form the basis of our perceptual space. Crucially, the filter is optimized jointly with the main forecasting model, ensuring that the learned perceptual space is explicitly oriented towards the prediction task. Finally, losses are calculated within these perception spaces to optimize the model. Code is available at \url{https://github.com/zyh16143998882/CP_Loss}.
\end{abstract}
\section{Introduction}
\label{sec:intro}

\begin{figure}[t]
    \begin{center}
    \includegraphics[width=\linewidth]{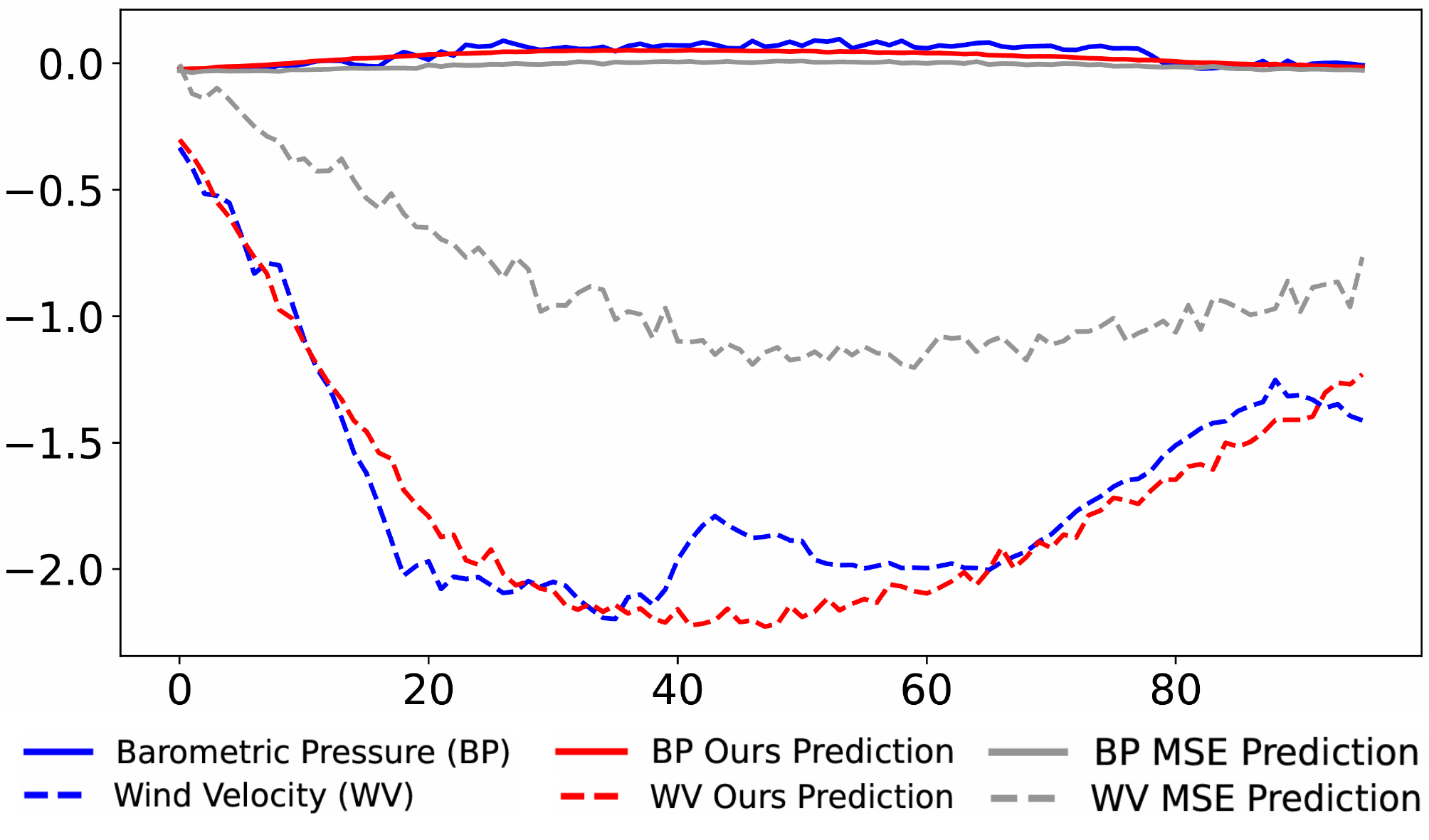}
    \caption{The prediction results of different channels.
    }\label{cur1}
    \end{center}
    \vspace{-1.0cm}
\end{figure}

Time-series data is a fundamental data type with widespread applications in critical domains such as financial markets \cite{financial1}, weather forecasting \cite{weather1,weather2}, and industrial automation \cite{casl,lcm}. In real-world scenarios, such data often manifests as multi-channel (or multivariate) sequences; for instance, a meteorological sensor array might simultaneously record variables like wind speed, precipitation, and barometric pressure. However, as shown in Fig. \ref{cur1}, these parallel channels exhibit significant heterogeneity in their dynamic properties: some channels may contain high-range, sharp fluctuations (e.g., wind velocity), while others are relatively smooth and slowly varying (e.g., barometric pressure). This inherent inter-channel disparity poses a significant challenge for building a predictive model that can capture the dynamics of all channels with equal fidelity.

To address this challenge, the dominant paradigm in current research focuses on designing increasingly complex deep learning architectures \cite{patchtst,periodicity,dlinear,amd,itransformer}. These models aim to implicitly learn diverse dynamics by capturing intricate intra- and inter-channel dependencies. During training, however, they are almost universally guided by a simple, generic loss function, such as Mean Squared Error (MSE). This reveals a fundamental contradiction: we are attempting to evaluate a model designed to capture heterogeneity using a "one-size-fits-all" metric. The limitation of such generic loss functions lies in their channel-agnostic nature, as they impose a uniform penalty on prediction errors across all channels, neglecting the data's inherent heterogeneity. Consequently, to minimize the total error, the model often compromises by catering to channels with larger numerical ranges or smoother variations. This leads to overly smoothed predictions for more volatile yet equally important channels, thereby losing crucial dynamic details.

This issue is conceptually analogous to a classic problem in image assessment where MSE is similarly inadequate for measuring human-perceived image quality. Extensive research \cite{ssim,msssim} has confirmed that the Human Visual System (HVS) is not equally sensitive to all pixel-wise differences; instead, it relies on high-level features such as structure, luminance, and contrast for a comprehensive judgment. To address this, researchers \cite{pl1,pl2} proposed computing errors by transforming images into a perceptually uniform space, an approach that has proven far superior to traditional pixel-level losses. We argue that time-series forecasting also requires a similar paradigm shift from superficial numerical matching to a deeper dynamics-aware \cite{idpt} approach.

To this end, we propose a novel \textbf{C}hannel-wise \textbf{P}erceptual Loss (\textbf{CP Loss}). Its core idea is to dynamically construct a dedicated perceptual space for each time-series channel, wherein the complex dynamics of the signal are effectively decomposed, rendering the error metric more exact. Specifically, we design a learnable, channel-wise filter based on redundancy reduction. This filter decomposes the raw signal into a set of disentangled, multi-scale representations, which collectively form the tailored perceptual space for that channel. To ensure this space is precisely aligned with the prediction task, the filter operates as a differentiable module within the loss function and is optimized jointly with the main prediction model. By calculating the prediction error within each channel's unique perceptual space, we provide the model with highly differentiated and precise gradient guidance, driving it to achieve more precise predictions.

\section{THE PROPOSED METHOD}
\label{sec:format}

\subsection{Related Work}

Traditionally, loss functions for time series forecasting centered on point-wise metrics like Mean Squared Error (MSE), which suffer from well-documented limitations such as producing over-smoothed forecasts. In response, a new generation of more sophisticated loss functions has emerged, broadly classified into two paradigms. The first, shape-focused losses, includes methods ranging from Dynamic Time Warping \cite{dtw}, TILDE-Q \cite{tilde} for assessing temporal alignment to more recent patch-wise structural loss \cite{psloss} that compares the properties of local segments. The second, dependency-focused losses, targets temporal correlations, such as FreDF \cite{fredf} operating in the frequency domain. 

Despite their advancements, these emerging paradigms still share two critical limitations that motivate our work. \textbf{First}, they remain predominantly channel-agnostic, applying a uniform similarity metric across all channels of a multivariate series, thereby ignoring their inherent heterogeneity. \textbf{Second}, they rely on pre-defined, fixed measures of similarity, such as a fixed set of statistical calculations or a standard Fourier transform. These static measures cannot adapt to the unique characteristics of a specific dataset or task.

Our proposed CP Loss is explicitly designed to overcome these limitations. First, to counter the channel-agnostic problem, CP Loss operates on a channel-wise basis, constructing a unique perceptual space tailored to each individual channel's dynamics. Second, to move beyond fixed similarity measures, CP Loss's perceptual space is not static. Instead, it is dynamically generated by a learnable, differentiable filter that serves as an integral component of the loss function itself. This filter is optimized jointly with the main forecasting model, which ensures the learned perceptual space is explicitly task-oriented. By calculating the loss within this adaptive, learned space, CP Loss achieves a fine-grained error measurement that is inherently tailored to data.

\subsection{Preliminaries}

Let $\mathbf{X} = \{\mathbf{x}_1, \dots, \mathbf{x}_{M}\} \in \mathbb{R}^{C \times M}$ be a multi-channel time series, where $C$ is the number of channels and $M$ is the lookback window length. The objective of time series forecasting is to learn a mapping function $g: \mathbb{R}^{C \times M} \to \mathbb{R}^{C \times N}$ that generates a prediction $\hat{\mathbf{Y}} = \{\hat{\mathbf{y}}_1, \dots, \hat{\mathbf{y}}_{N}\} \in \mathbb{R}^{C \times N}$ to approximate the ground truth future sequence $\mathbf{Y} = \{\mathbf{y}_1, \dots, \mathbf{y}_{N}\} \in \mathbb{R}^{C \times N}$ over a forecast horizon $N$. This learning process is guided by minimizing a loss function that measures the discrepancy between the prediction and the ground truth. A dominant choice is the Mean Squared Error, which computes the average squared difference for each channel $c$ and then averages the results. To obtain the final loss $\mathcal{L}$, the losses from all channels are aggregated:
\begin{equation} \label{eq:mse_total}
    \mathcal{L} = \frac{1}{C \cdot N} \sum_{c=1}^{C} \sum_{n=1}^{N} (y_{n}^{(c)} - \hat{y}_{n}^{(c)})^2.
\end{equation}

\subsection{Channel-wise Perceptual Loss}

\begin{figure}[t]
    \begin{center}
    \includegraphics[width=\linewidth]{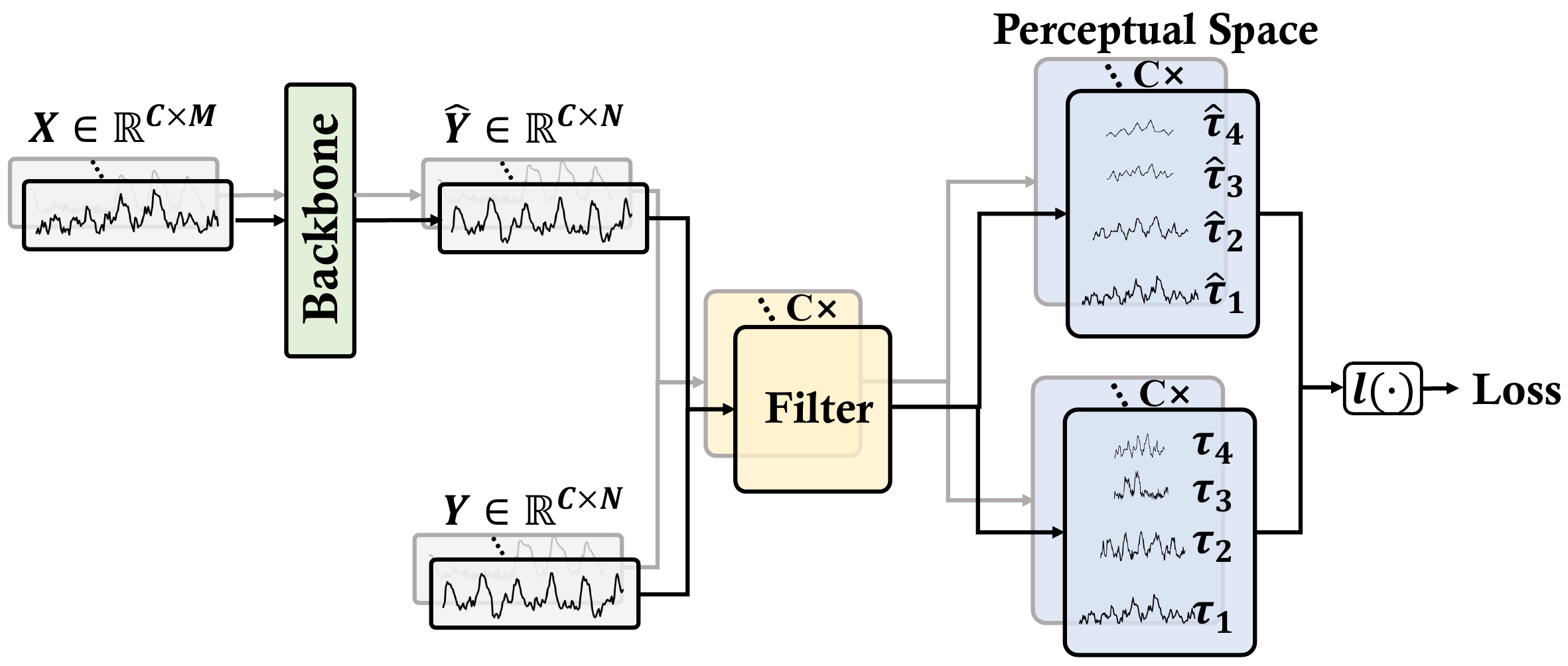}
    \caption{The pipeline of our channel-wise perceptual loss.
    }\label{frame}
    \end{center}
    \vspace{-0.5cm}
\end{figure}

Our proposed method, the Channel-wise Perceptual Loss (CP Loss), is designed to guide forecasting models by measuring errors in a dynamically learned perceptual space rather than in the raw signal space. The overall pipeline, illustrated in Fig.~\ref{frame}, consists of two primary stages: (1) a standard backbone model for generating predictions, and (2) a perceptual loss module that transforms both predictions and ground truth into a perceptual space sequence for comparison.

\subsubsection{Perceptual Space Transformation}
The core of our method is the transformation from the raw signal space to a learned perceptual space. This process begins after a backbone model $g$ generates a prediction $\hat{\mathbf{Y}} = g(\mathbf{X})$. Instead of directly comparing $\hat{\mathbf{Y}}$ with the ground truth $\mathbf{Y}$, both sequences are fed in parallel into a learnable, channel-wise filter, denoted as $f_{\theta}= \{f_{\theta}^1, \dots, f_{\theta}^{C}\}$. This filter must share the same set of weights $\theta$ when processing both the prediction and the ground truth. The filter $f_{\theta}$ is designed to decompose the time series signal for each channel into a set of $K$ disentangled, multi-scale representations based on the idea of reducing redundancy. For a given input sequence $\mathbf{Y}$, the transformation is:
\begin{equation}
    \{\mathbf{\tau}_1, \dots, \mathbf{\tau}_K\} = f_{\theta}(\mathbf{Y}).
\end{equation}
This collection of representations $\{\mathbf{\tau}_k\}_{k=1}^K$ collectively constitutes the perceptual space.

\subsubsection{Loss Computation and Joint Optimization}
With the prediction and ground truth transformed into their respective perceptual representations, $\{\hat{\mathbf{\tau}}_k\}_{k=1}^K$ and $\{\mathbf{\tau}_k\}_{k=1}^K$, the loss is computed by aggregating the discrepancies between corresponding components. A base distance metric $l(\cdot, \cdot)$, such as Mean Absolute Error in our implementation, is applied to each pair of representations. The final Channel-wise Perceptual Loss is the sum of these component-wise losses:
\begin{equation}
    \mathcal{L}_{\text{CP Loss}} = \sum_{k=1}^{K} l(\mathbf{\tau}_k, \hat{\mathbf{\tau}}_k).
\end{equation}
The entire system, including the parameters of the backbone model $g$ and the parameters $\theta$ of our filter $f_{\theta}$, is optimized in a joint fashion. This training paradigm is crucial, as it compels the filter to learn representations that are explicitly useful for the forecasting task. Consequently, the perceptual space is dynamically tailored to the specific characteristics of the data and the objective of minimizing prediction error.

\subsubsection{The Detail of Perceptual Filter}

\begin{figure}[t]
    \begin{center}
    \includegraphics[width=\linewidth]{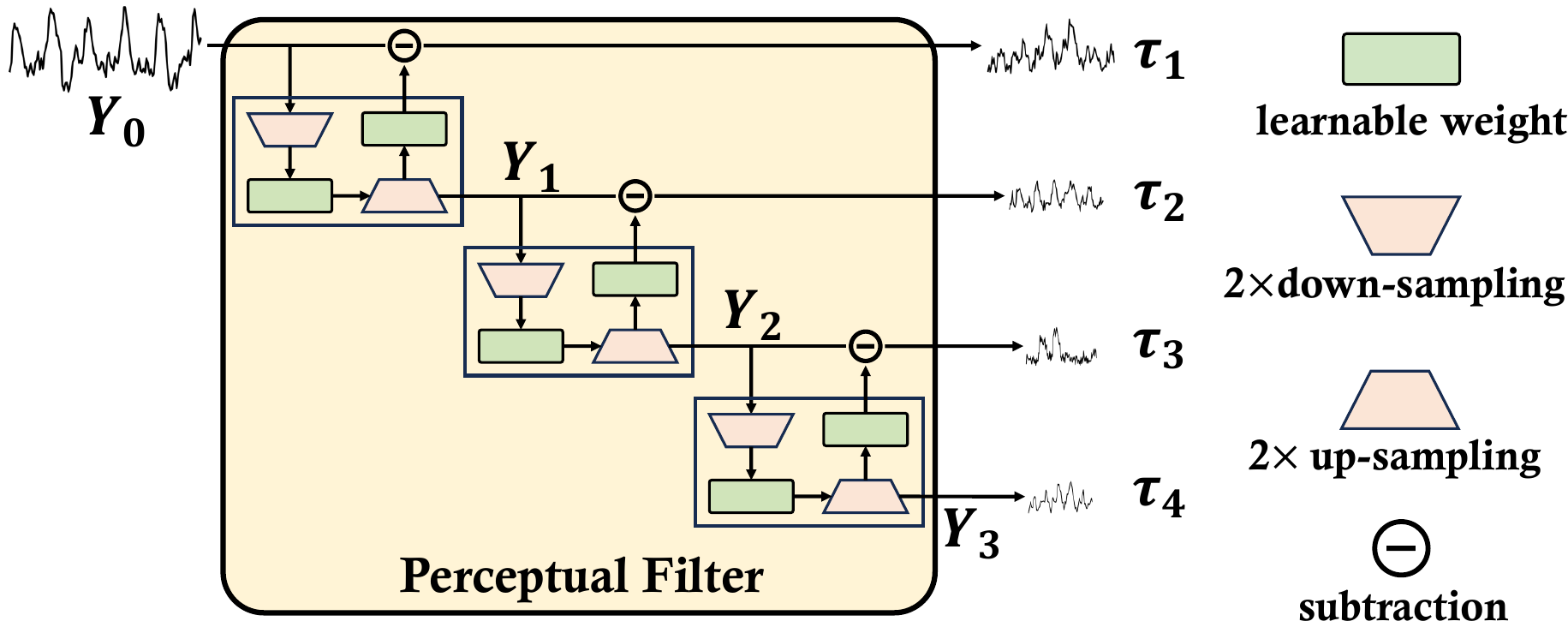}
    \caption{The detail of our perceptual filter.
    }\label{filter}
    \end{center}
\end{figure}

The core of our method is the Perceptual Filter, $f_{\theta}$, which is responsible for transforming a raw time series into its multi-scale perceptual representations. As illustrated in Fig.~\ref{filter}, our filter operates on the principle of hierarchical redundancy reduction, utilizing learnable weights to perform a representation decomposition explicitly tailored to the forecasting task.

The decomposition process is hierarchical. Let $\mathbf{Y}_0$ be the original input time series for a single channel. At the first level, a smoothed-and-reconstructed version of the signal is generated. This is achieved by first passing $\mathbf{Y}_0$ through a $2\times$ down-sampling operation, followed by a transformation with a learnable weight (e.g., a 1D convolution), and then a $2\times$ up-sampling operation to restore the original temporal resolution. The resulting sequence, which can be seen as a low-pass filtered version of the original, is then subtracted from $\mathbf{Y}_0$. This subtraction isolates the high-frequency details, which form our first perceptual representation, $\mathbf{\tau}_1$. The down-sampled, learnable transformation of $\mathbf{Y}_0$ serves as the input approximation, $\mathbf{Y}_1$, for the next level. This process is repeated iteratively: at each level $k$, the detail component $\mathbf{\tau}_k$ is computed by subtracting a reconstructed version of the approximation $\mathbf{Y}_{k-1}$, and the new, coarser approximation $\mathbf{Y}_k$ is passed down to the subsequent level.

The output of the Perceptual Filter is the complete set of decomposed representations. In the depicted three-level architecture, the outputs are $\{\mathbf{\tau}_1, \mathbf{\tau}_2, \mathbf{\tau}_3, \mathbf{\tau}_4\}$. The components $\mathbf{\tau}_1$ through $\mathbf{\tau}_3$ are the detail representations, capturing information at progressively coarser scales. Specifically, $\mathbf{\tau}_1$ contains the finest, highest-frequency patterns, while $\mathbf{\tau}_3$ contains lower-frequency variations. The final component, $\mathbf{\tau}_4$, is the coarsest approximation of the signal remaining after all details have been extracted, typically representing the signal's fundamental trend or low-frequency base. By optimizing the filter's learnable weights jointly with the forecasting model, our framework ensures that this decomposition is not arbitrary but is instead tailored to extract the structural features most salient for accurate prediction.

\section{Experiments}

Our experiments were performed on six real-world multivariate time series datasets: the four ETT variants (ETTh1, ETTh2, ETTm1, ETTm2), Weather, and ECL. All data follow the standard protocol
used by previous works \cite{patchtst,itransformer}.

We select four leading time series forecasting models with diverse architectures to comprehensively evaluate the CP loss function. Specifically, we include Transformer-based models: iTransformer \cite{itransformer}; MLP-based models: DLinear \cite{dlinear}, TimeMixer \cite{timemixer}; and CNN-based models: TimesNet \cite{timesnet}.

\begin{table*}[htbp]
  \centering
  \resizebox{\textwidth}{!}{
    \begin{tabular}{c|cccc|cccc|cccc|cccc|}
    \toprule
    Models & \multicolumn{4}{c|}{iTransformer} & \multicolumn{4}{c|}{TimeMixer} & \multicolumn{4}{c|}{Dlinear}  & \multicolumn{4}{c|}{TimseNet} \\
    \midrule
    Loss  & \multicolumn{2}{c}{Vanilla} & \multicolumn{2}{c|}{CP Loss} & \multicolumn{2}{c}{Vanilla} & \multicolumn{2}{c|}{CP Loss} & \multicolumn{2}{c}{Vanilla} & \multicolumn{2}{c|}{CP Loss} & \multicolumn{2}{c}{Vanilla} & \multicolumn{2}{c|}{CP Loss} \\
    \midrule
    Metric & MSE   & MAE   & MSE   & MAE   & MSE   & MAE   & MSE   & MAE   & MSE   & MAE   & MSE   & MAE   & MSE   & MAE   & MSE   & MAE \\
    \midrule
    ETTh1 & 0.457 & 0.449 & \textbf{0.447} & \textbf{0.438} & 0.460 & 0.444  & \textbf{0.452} & \textbf{0.430} & 0.445 & 0.454 & \textbf{0.417} & \textbf{0.425} & 0.457 & 0.460  & \textbf{0.447} & \textbf{0.439} \\
    ETTh2 & 0.384 & 0.407 & \textbf{0.378} & \textbf{0.400} & 0.392 & 0.410  & \textbf{0.378} & \textbf{0.399} & 0.469 & 0.463 & \textbf{0.408} & \textbf{0.421} & 0.406 & 0.420  & \textbf{0.385} & \textbf{0.403} \\
    ETTm1 & 0.408 & 0.412 & \textbf{0.397} & \textbf{0.393} & 0.388 & 0.399 & \textbf{0.381} & \textbf{0.382} & 0.359 & 0.381 & \textbf{0.354} & \textbf{0.370} & 0.417 & 0.421 & \textbf{0.394} & \textbf{0.397} \\
    ETTm2 & 0.292 & 0.336 & \textbf{0.284} & \textbf{0.323} & 0.278 & 0.326 & \textbf{0.274} & \textbf{0.317} & 0.283 & 0.345 & \textbf{0.258} & \textbf{0.312} & 0.299 & 0.334 & \textbf{0.291} & \textbf{0.325} \\
    Weather & 0.261 & 0.281 & \textbf{0.254} & \textbf{0.271} & 0.245 & 0.276 & \textbf{0.245} & \textbf{0.265} & 0.247 & 0.300   & \textbf{0.244} & \textbf{0.277} & 0.259 & 0.285 & \textbf{0.253} & \textbf{0.275} \\
    ECL   & 0.176 & 0.267 & \textbf{0.175} & \textbf{0.261} & 0.182 & 0.272 & \textbf{0.182} & \textbf{0.271} & 0.167 & 0.264 & \textbf{0.167} & \textbf{0.261} & 0.196 & 0.297 & \textbf{0.194} & \textbf{0.293} \\
    \bottomrule
    \end{tabular}%
  }
  \caption{Long-term multivariate forecasting results. The table reports MSE and MAE for different forecasting lengths.}
  \label{res1}%
\end{table*}%

\subsection{Main Results}

The comprehensive results, summarized in Table \ref{res1}, demonstrate the consistent and significant benefits of our Channel-wise Perceptual Loss. We report the average MSE and MAE across four challenging prediction horizons {96, 192, 336, 720} on six real-world benchmarks. The findings are clear: when integrated with our CP Loss, every backbone model (iTransformer, TimeMixer, DLinear, and TimesNet) achieves a notable performance improvement over the standard MSE baseline on every dataset. This universal enhancement underscores the broad applicability and effectiveness of our proposed loss function.

\subsection{Comparison with Other Loss Functions}

\begin{table}[t]
  \centering
  \resizebox{\linewidth}{!}{
    \begin{tabular}{c|c|ccccc}
    \toprule
    \multicolumn{2}{c}{Dataset} & MSE   & TILDE-Q & FreDF & PS Loss & CP Loss \\
    \midrule
    \multirow{5}[4]{*}{ETTm1} & 96    & 0.377 & 0.366 & 0.367 & 0.360  & \textbf{0.351} \\
          & 192   & 0.396 & 0.392 & 0.393 & 0.384 & \textbf{0.377} \\
          & 336   & 0.418 & 0.416 & 0.414 & 0.406 & \textbf{0.403} \\
          & 720   & 0.457 & 0.453 & 0.451 & \textbf{0.440} & 0.441 \\
\cmidrule{2-7}          & Avg   & 0.412 & 0.407 & 0.406 & 0.398 & \textbf{0.393} \\
    \midrule
    \multirow{5}[4]{*}{Weather} & 96    & 0.216 & 0.208 & 0.210  & 0.203 & \textbf{0.201} \\
          & 192   & 0.260  & 0.252 & 0.254 & 0.249 & \textbf{0.248} \\
          & 336   & 0.299 & 0.296 & 0.297 & \textbf{0.291} & \textbf{0.291} \\
          & 720   & 0.348 & 0.346 & 0.348 & \textbf{0.343} & 0.344 \\
\cmidrule{2-7}          & Avg   & 0.281 & 0.276 & 0.277 & 0.272 & \textbf{0.271} \\
    \bottomrule
    \end{tabular}%
  }
  \caption{Comparison MAE metric between the proposed channel-wise perceptual loss and other loss functions.}
  \label{res2}%
\end{table}%

As shown in Table \ref{res2}, our CP Loss demonstrates superior overall performance when benchmarked against other state-of-the-art loss functions, including the shape-based TILDE-Q, PS Loss, and frequency-based FreDF. Our results achieved the lowest average MAE on both the ETTm1 and Weather datasets. While PS Loss is highly competitive, especially at the longest prediction horizons, the consistently higher performance of CP Loss validates the power of its learnable, task-oriented approach over methods with fixed similarity metrics.

\subsection{More Analysis}

\textbf{Computational complexity.} The proposed CP Loss is exceptionally efficient in both its parameter count and computational load. The number of learnable parameters is minimal, determined solely by a single learnable filter with a size of $C \times k$, where $C$ is the number of channels and $k$ is a small, fixed kernel size, and we set $k=5$ in our implementation. This results in a parameter count that scales linearly with $C$. Computationally, the module's overhead is dominated by a series of 1D convolutions within its multi-level structure, leading to a complexity that scales linearly with the batch size ($B$), number of channels ($C$), and prediction length ($N$). This dual efficiency in parameters and computation establishes the CP Loss as a lightweight and scalable component.

\begin{figure}[t]
    \begin{center}
    \includegraphics[width=\linewidth]{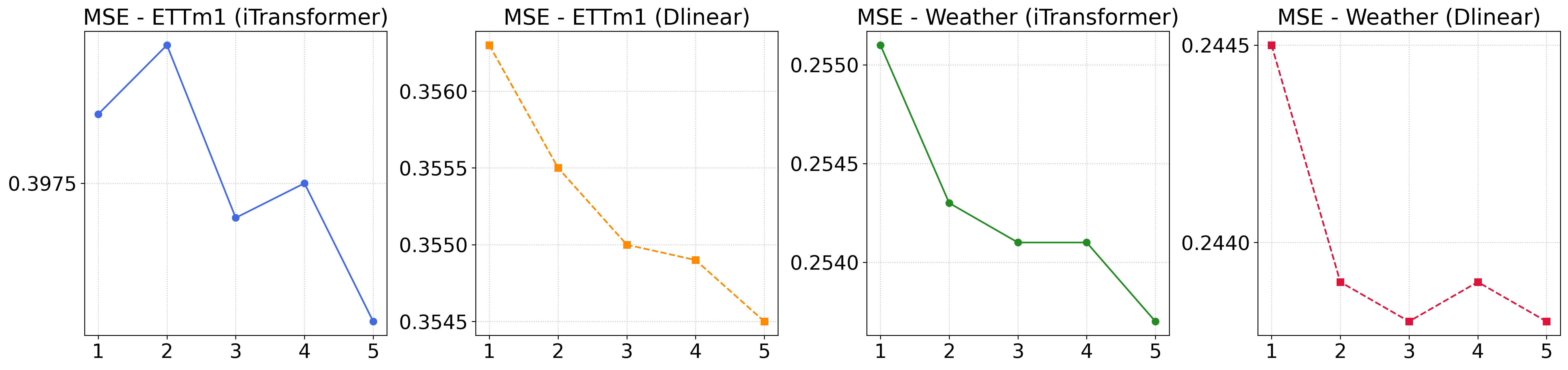}
    \caption{The effect of different scales.
    }\label{cur}
    \end{center}
\end{figure}

\textbf{The effect of different scales.} To understand the impact of the decomposition depth, we performed an ablation study on the number of scales from 1 to 5. We evaluated performance on two datasets and backbone architectures. As shown in Fig. \ref{cur}, there is a clear trend: prediction error decreases as we set more decomposition levels. This suggests that a deeper, multi-scale analysis allows the model to capture a richer set of temporal dynamics. Across our experiments, we observed that the optimal performance was achieved at a scale of 5, which we subsequently used for our main results.

\section{Conclusion}

In this work, we introduced the Channel-wise Perceptual Loss, a novel loss function designed to overcome the limitations of channel-agnostic metrics in multivariate forecasting. Our method employs a learnable, channel-wise filter to dynamically construct a unique perceptual space for each time series. By decomposing the signal into disentangled, multi-scale representations, our loss provides a more precise gradient than traditional losses.

\section{Acknowledgements}

This work is supported in part by the National Natural Science Foundation of China, under Grant (62302309,62571298).

\clearpage
\bibliographystyle{IEEEbib}
\bibliography{strings,refs}

\end{document}